\documentclass[sensors,article,accept,moreauthors,pdftex]{Definitions/mdpi} 
\usepackage{arydshln}

\firstpage{1} 
\pubvolume{xx}
\issuenum{1}
\articlenumber{5}
\pubyear{2020}
\copyrightyear{2020}
\history{Received: 20 March 2020; Accepted: 28 April 2020 ; Published: date}
\updates{yes} 
\usepackage{graphicx}
\usepackage{pgfplots}
\usepackage{mathrsfs}
\usepackage{algorithmic}
\usepackage[lined,boxed,ruled,commentsnumbered]{algorithm2e}
\makeatletter
\AtBeginDocument{\let\hl\@firstofone}
\makeatother

\Title{Recognition of Patient Groups with Sleep Related Disorders using Bio-signal Processing and Deep~Learning 
}%

\Author{\textls[-5]{\hl{Delaram Jarchi} $^{1,2,}$*\orcidA{}, \hl{Javier Andreu-Perez} $^{1,2}$\orcidB{}, Mehrin Kiani $^{1}$, Oldrich Vysata $^{3,4}$, Jiri Kuchynka~$^{4}$, Ales Prochazka $^{3,5}$ and Saeid Sanei $^{6}$}}
\AuthorNames{Firstname Lastname, Firstname Lastname and Firstname Lastname}

\address{%
$^{1}$ \quad Smart Health Technologies Group, School of Computer Science and Electronic Engineering; University of Essex, \hl{Colchester  CO4 3SQ, UK}; delaram.jarchi@essex.ac.uk \hl{(D.J.)}; javier.andreu@essex.ac.uk (J.A.P.); mehrin.kiani@essex.ac.uk  (M.K.)\\
$^{2}$ \quad Embedded and Intelligent Systems Laboratory, School of Computer Science and Electronics, University of Essex, \hl{Colchester  CO4 3SQ, UK}\\
$^{3}$ \quad Department of Computing and Control Engineering, University of Chemistry and Technology in Prague,
166 28 Prague 6, Czech Republic; Oldrich.Vysata@fnhk.cz (O.V.);  A.Prochazka@ieee.org (A.P.)
\\
$^{4}$ \quad Department of Neurology, Faculty of Medicine in Hradec Králové, Charles University,
\mbox{500 05 Hradec Králové, Czech Republic}; Jiri.Kuchynka@fnhk.cz {\hl{(J.K.)}}
\\
$^{5}$ \quad Czech Institute of Informatics, Robotics and Cybernetics, Czech Technical University in Prague, \mbox{160 00 Prague 6, Czech Republic}
\\
$^{6}$ \quad School of Science and Technology, Nottingham Trent University, \hl{Nottingham NG11 8NS, UK}; saeid.sanei@ntu.ac.uk \hl{(S.S.)}}

\corres{\hangafter=1 \hangindent=1.05em \hspace{-0.82em}Correspondence: delaram.jarchi@essex.ac.uk}

\abstract{Accurately diagnosing sleep disorders is essential for clinical assessments and treatments. Polysomnography (PSG) has long been used for detection of various sleep disorders. In this research, electrocardiography (ECG) and electromayography (EMG) have been used for recognition of breathing and movement-related sleep {disorders}. Bio-signal processing has been performed by extracting EMG features exploiting entropy and statistical moments, in addition to developing an iterative pulse peak detection algorithm using synchrosqueezed wavelet transform (SSWT) for reliable extraction of heart rate and breathing-related features from ECG. A deep learning framework has been designed to incorporate EMG and ECG features. The~framework has been used to classify four groups: healthy subjects, patients with obstructive sleep apnea (OSA), patients with restless leg syndrome (RLS) and patients with both OSA and RLS. The~proposed deep learning framework produced a mean accuracy of {72\%} and weighted F1 score of {0.57} across subjects for our formulated four-class problem.}


\begin{document}

\section{Introduction}\label{s1}
Study of human sleep and the underlying physiological changes is crucial for detecting sleep disorders, improving the quality of sleep and avoiding daytime sleepiness. There are two main sleep disorders, including obstructive sleep apnea (OSA) \cite{apnea,apnea1} as a breathing-related sleep disorder, and restless leg syndrome (RLS) as a movement-related sleep disorder~\cite{leg1,rest,leg}.
OSA is associated with full or partial occlusion of the upper airway during sleep which needs treatment; otherwise, it can affect the cardiovascular system and also lead to nocturnal death~\cite{osa,osa1}.
On the other hand, RLS is associated with uncontrollable sporadic or periodic leg movements during~sleep.

Polysomnography (PSG) {as a type of sleep study} has been developed for noninvasive analysis of sleep. It includes a number of wearable sensors, such as an electrocardiogram (ECG) {to record the cardiac activity from electrodes attached to chest}; a {pulse oximeter attached to a finger to} record photoplethysmography (PPG) {for blood oxygen saturation measurement}; an electromyograph (EMG) that can be fitted to the chin and leg { for movement analysis}; an electroencephalograph (EEG) {to record brain activity that is useful for sleep stage classification}; an electrooculograph (EOG) {to monitor eye movements}; a nose sensor {for monitoring nasal breathing}; and an accelerometer {to analyze chest movements}. \textls[-15]{PSG~has been used for detecting various sleep disorders and also measurement of sleep quality as the gold standard.}

Many techniques have been developed to detect sleep stages and sleep disorders, such as OSA using PSG in research studies and clinical trials. 
{
EEG has been used to classify sleep stages 
using correlation graphs~\cite{abdula} and weighted undirected complex networks~\cite{dikh}.
EEG has also been used to detect sleep apnea events using inter-band energy ratio
~\cite{saha}. 
In another study, EEG and EMG signals were used to detect sleep apnea exploiting EEG arousal~\cite{sugi}.
A combination of EEG, ECG and EMG has been used in~\cite{moridani} to detect sleep apnea events in different sleep stages.
A non-contact pressure-sensitive device was built \cite{mosq} to detect OSA.
\hl{Deriving breathing patterns from a single channel ECG is more straightforward than analysis of EEGs, which are prone to artifacts and drifts~{\cite{eega}}, and usually more than one EEG channel is required for apnea detection}. The~benefit of using ECG in our framework is to further motivate the development and adaptation of similar techniques in future studies for PPG analysis from wrist worn optical sensors to make unobtrusive and portable systems for identification of sleep disorders.}

Reducing the number of wearable sensors in the PSG system for detection of sleep disorders alleviates the system complexity and focuses on the required information only. 
Single-lead ECG has been successfully used to identify OSA~\cite{bsoul,jarchi,jarchi1}. The~key in analysis of OSA using ECG is to accurately extract the breathing signal, since OSA is directly related to respiration. 
All the techniques developed for ECG analysis can be simply adapted for PPG analysis which is a very non-intrusive way of measuring human physiological~parameters. 

{EMG has been used for detection of neuromuscular diseases which can arise from sleep movement disorders~\cite{shokr}.} 
EMG has {also} been shown to be particularly useful in the detection of RLS. There is evidence that RLS and OSA have been associated as there are more periodic leg movements during sleep apnea episodes~\cite{rest1}. However, concurrent and quantitative analysis of RLS and OSA have not been well~studied. 

The objective of this paper is to exploit strong signal processing and machine learning techniques for joint identification of OSA and RLS using ECG and EMG. This will open a new insight into identification and correlation analysis of respiratory and movement-related symptoms. More specifically, a~deep-learning-based framework is proposed for automatic identification and classification of healthy subjects, patients with OSA, patients with RLS and patients with both OSA and RLS. This is also useful for group-specific sleep analysis of sleep stages and scoring~\cite{ales,ales1}. 

\textls[-15]{The remainder of the paper is as follows. In~Section~\ref{s2}, first the EMG analysis is introduced which includes description of features such as lower and higher-order statistics and a relatively new entropy-related measure called {dispersion entropy (DE)} to incorporate signal fluctuations~\cite{azami}. This is followed by} the ECG analysis subsection. In~this subsection, first a strong signal processing technique for time-frequency analysis, i.e.,~synchrosqueezed wavelet transform (SSWT), is described. Then, an~iterative pulse peak detection technique which uses SSWT as a core technique is introduced for reliable ECG pulse peak detection. Then, ECG derived features, including heart rate variability and respiratory-related measures are described. At~the end of this section, a~deep learning framework is introduced which exploits ECG and EMG features for classification of four groups, including healthy subjects, patients with OSA, patients with RLS and patients with both OSA and RLS. In~Section~\ref{s3}, the~results are provided. Finally, Section~\ref{s4} concludes the paper, highlighting the major contributions, limitations and potential improvements of the proposed framework for future sleep analysis~systems. 


\unskip

\section{Materials and~Methods\label{s2}}
\unskip
\subsection{EMG~Analysis}
In order to characterize the distribution of EMG data samples, quantitative measures should be used. \textls[-15]{In~the following, lower and higher-order statistics and entropy-based measures which are useful to
quantify changes in the EMG signals are described. 
In those cases where EMG is contaminated by ECG, it~can be
effectively restored using the method based on singular spectrum analysis proposed in~\cite{sanei}.}

\subsubsection{EMG Features from~Moments}

 The first moment is simply the data mean and denoted as $E[X]$, where $X$ {is considered as a real-valued random variable}. The~second moment is the variance; its square root can be calculated using $(E[(X-E[{X}])^2])^\frac{1}{2}$ as the standard deviation. These two moments are considered low-order statistics. Similarly, higher order statistics {including} skewness and kurtosis {are derived using the following equations:}
\begin{equation}
    Skewness = \frac{E[{(X-E[{X}])}^3]}{(E[(X-E[{X}])^2])^\frac{3}{2}} 
\end{equation}
\begin{equation}
    Kurtosis = \frac{E[{(X-E[{X}])}^4]}{(E[(X-E[{X}])^2])^2} 
\end{equation}
{where $X$ is a real-valued random variable. Both lower and higher order statistics can be applied to EMG signals for feature extraction}. 

\subsubsection{EMG Features from Entropy {Measurement}}
Entropy has been used to quantify the signal uncertainty and dynamical characteristics. Sample entropy~\cite{se} and permutation entropy~\cite{pe} have been widely used to indicate the entropy measure. 
Here, we have selected a recently developed entropy-based method called DE which takes into account the signal fluctuations~\cite{azami}. 
The algorithm to estimate the DE is summarized in four main steps below:

\textit{\hl{Step One}}: The {samples} ${x}_j(j=1,2,\dots,N)$ of signal $\mathbf{x}$ are mapped into $c$ classes that are labeled from 1 to $c$ {($c$ is a user-defined parameter)}. As~proposed in~\cite{azami}, normal cumulative distribution function (NCDF) is used to map $\mathbf{x} {= \{{x}_1,{x}_2,\dots,{x}_N\}}$ into $\mathbf{y} = \{{y}_1,{y}_2,\dots,{y}_N\}$, {from 0 to 1, first}. Then, a~linear algorithm  is applied to assign each  ${y}_j$ value to an integer from 1 to $c$ ($z^{c}_{j}=round(c.y_j+0.5)$ where $z^{c}_{j}$ is the $j$th member of the classified time-series).

\textit{\hl{Step Two}}: {E}mbedding vectors with embedding dimension $m$ and time delay $d$ {are} created based on: $z_{i}^{m,c}=\{z^{c}_{i},z^{c}_{i+d},\dots,z^{c}_{i+(m-1)d}\}, i = 1,2,\dots,N-(m-1)d$. Then, each produced time series 
$z_{i}^{m,c}$ is mapped to a dispersion pattern $\pi_{v_0v_1\dots v_{m-1}}$, where $z_{i}^{c}=v_0, z_{i+d}^{c}=v_1,\dots,z_{i+(m-1)d}^{c}=v_{m-1}$. 
{The~number of potential dispersion patterns that can be assigned to each time series ($z_{i}^{m,c}$) is equal to ${c^m}$. This is due to the fact that the time-series contains $m$ members and each member can be assigned to an integer from 1 to $c$.}

\textit{\hl{Step Three}}:
For each possible dispersion pattern (there are ${c^m}$ potential patterns), a~relative frequency is calculated as:
\begin{equation}
    p(\pi_{v_0,v_{1},\dots, v_{m-1}})=\frac{Number\{i|i{\leq N}-(m-1)d,z_{i}^{m,c} ~\text{has type}~ \pi_{v_0v_1\dots v_{m-1}} \}}{N-(m-1)d}
\end{equation}

Based on the above equation, $p(\pi_{v_0,v_{1},\dots, v_{m-1}})$ demonstrates the number of dispersion patterns $\pi_{v_0v_1\dots v_{m-1}}$ which are assigned to $z_{i}^{m,c}$, divided by the total number of embedded time-series (embedding dimension is $m$).

\textit{\hl{Step Four}}: The DE value can be calculated using the Shannon's definition of entropy as~\cite{shannon}:
\begin{equation}
DE(\mathbf{x},m,c,d)=-\sum_{\pi=1}^{c^m}p(\pi_{v_0,v_{1},\dots, v_{m-1}}).ln(p(\pi_{v_0,v_{1},\dots, v_{m-1}}))
\end{equation}

A summary of EMG features including moments and DE is provided in Table~\ref{table:feat}.

\begin{table}[H]
\caption{A summary of features extracted from \hl{EMG} and \hl{ECG} signals is provided~below.}
\label{table:feat}
\centering
\scalebox{0.85}[0.85]{

\begin{tabular}{cc}

\toprule
\textbf{~~~~~~~~~~~~~~~~~~~~~~~~~~~~~~~~~~~~~~~~~~~~~~~~~~~~~~~~~~~~~~~~~~~~~~~~~~~~~~~~~~~~~~~~~~{EMG Features}}\\
\midrule
\textbf{{Feature}}&\textbf{{Feature type$^{*}$}}	\\
\midrule 
Mean &	{Raw}		\\
Standard deviation	&	{Raw}		\\
Skewness&	{Raw}			\\
Kurtosis&	{Raw}		\\
Dispersion Entropy	&	{Raw}		\\
\midrule 
\textbf{~~~~~~~~~~~~~~~~~~~~~~~~~~~~~~~~~~~~~~~~~~~~~~~~~~~~~~~~~~~~~~~~~~~~~~~~~~~~~~~~~~~~~~~~~~{ECG Features}}	\\
\midrule
\textbf{{Feature}}&\textbf{{Feature type}}	\\
\midrule 
Maximum difference {of pulse peaks} &{Time-domain}			\\
Minimum difference {of pulse peaks} &	{Time-domain}		\\
Mean difference	{of pulse peaks}	&{Time-domain}		\\
\midrule
Maximum amplitude { of respiratory amplitude modulation 	}		&{Time-domain}\\
Minimum amplitude {of respiratory amplitude modulation}			&{Time-domain}\\
Mean amplitude {of respiratory amplitude modulation	}		&{Time-domain}\\
\midrule
Maximum {of instantaneous frequencies of respiratory amplitude modulation}	&{Frequency-domain}\\
Minimum {of instantaneous frequencies of   respiratory amplitude modulation	}		&{Frequency-domain}\\
Mean {of instantaneous frequencies of  respiratory amplitude modulation	}
&{Frequency-domain}\\
{Standard deviation of instantaneous frequencies of  respiratory amplitude modulation	}		&
{Frequency-domain}\\
\bottomrule
\end{tabular}}\\
\begin{tabular}{cc}
\multicolumn{1}{c}{\footnotesize $^{*}$ Raw feature type means the features are directly extracted form signal samples.}
\end{tabular}

\end{table}

\subsection{ECG~Analysis}\label{s2.2}
In order to extract features from ECG, the~ECG pulse peaks (R peaks) should be detected. Although~many peak detection algorithms have been developed for ECG peak detection, the~reliability and generalisation of most of these algorithms have not been comprehensively evaluated. Here, we propose an iterative R peak detection algorithm which combines the temporal domain technique (simple pulse peak detection) with time-frequency spectral presentation of the signal. Such a hybrid technique that uses both temporal information of the ECG signal and time-frequency-based representation has significantly increased the reliability of pulse peak detection~algorithm.   

\subsubsection{Synchrosqueezed Wavelet~Transform}
SSWT applies the time-frequency analysis method to produce the spectrum of the input signal in the first stage. In~the second stage, a~reassignement technique is applied to the resultant time-frequency spectrum~\cite{sst}. The~objective of the reassignment technique is to enhance the initial spectrum to produce a sharper spectrum with visible and narrower frequency ridges. SSWT was first targeted for auditory signal analysis following its development~\cite{sst,sst1}. 
The SSWT algorithm is described in the following. \textit{\hl{Stage 1}}: The continuous wavelet transform (CWT) is applied to the input signal. Consider the CWT of the input signal $\mathbf{s}$ as:
\begin{equation}
 W_s(a,b) = \int_{-\infty}^{\infty} s(t)a^{-1/2}\overline{\psi(\frac{t-b}{a})} dt
\end{equation}
 where $a$ is the wavelet scale, $t$ is the time index, $\psi$ is the selected mother wavelet ($\overline{\psi}$ represents its complex conjugate form) and~$b$ is the position parameter. 
 A purely harmonic signal has been selected to better {explain} the SSWT algorithm. 
 {Suppose} that the mother wavelet is concentrated {on} the positive frequency {axis} such that $\hat{\psi}(\varepsilon)=0$ for $\varepsilon < 0$.
 The {selected harmonic} presented as $s(t)=A\cos(\omega t)$ {is} used in the following equations to demonstrate the operation of the SSWT. Based on the Plancherel's theorem~\cite{sst}, and~using the harmonic signal as the input, CWT can be formulated as:
\begin{equation}
\begin{split}
W_s(a,b) &= \int_{-\infty}^{\infty} s(t)a^{-1/2}\overline{\psi(\frac{t-b}{a})}dt\\
&= \frac{1}{2\pi} \int_{-\infty}^{\infty} \hat{s}(\varepsilon)a^{1/2}\overline{\hat{\psi}(a\varepsilon)}e^{ib\varepsilon}d\varepsilon\\
&= \frac{A}{4\pi} \int_{0}^{\infty} [\delta(\varepsilon-\omega)+\delta(\varepsilon+\omega)]a^{1/2}\overline{\hat{\psi}(a\varepsilon)}e^{ib\varepsilon}d\varepsilon \\
\end{split}
\end{equation}

{If} $\hat{\psi}(\varepsilon)$ is concentrated around $\varepsilon = \omega_{0}$, then $W_{s}(a,b)$ is concentrated around $a=\omega_{0}/\omega$, and~spreads out over a region of the horizontal line $a$:
\begin{equation}
 W_s(a,b) = \frac{A}{4\pi} a^{1/2}\overline{\hat{\psi}(a\omega)}e^{ib\omega}
\end{equation}

If $\omega=\omega_{0}/a$ is similar but not necessarily exactly identical to the actual instantaneous frequency (IF) of the input signal, some non-zero energy for $W_s(a,b)$ will appear. 
This energy needs to be moved away from $\omega$, and this is the main objective of the synchrosqueezing concept. 
Reassigning of the frequency locations nearer to the actual IF is a key to SSWT which enhances time-frequency spectral representation.    
\textit{Stage 2}: In the second stage, {for any (a,b)}, for~which $W_{s}(a,b)\neq0$, the~candidate IFs $(\omega_{s}(a,b))$ should be calculated:
\begin{equation}
\omega_{s}(a,b) = -i(W_{s}(a,b))^{-1}\frac{\partial}{\partial b}W_{s}(a,b)
\end{equation}

{For t}he selected signal which is purely harmonic signal $s(t)=A\cos(\omega t)$, $\omega_{s}(a,b)$ are simply derived as $\omega$ (the frequency of the harmonic signal) \cite{sst}. The~candidate IFs are used to recover actual frequencies. In~the synchrosqueezing step, re-allocation technique is used to map the time domain into the time-frequency domain (using $(b,a)\Rightarrow(b,\omega_{s}(a,b))$). 

{The discrete-frequency synchrosqueezed transform of $\mathbf{s}$} is defined as $T_{s}(\omega_{l},b)$:
\begin{equation}
 T_{s}(\omega_{l},b) = {(\Delta\omega)}^{-1}
 \sum_{a_k:|\omega_{s}(a_k,b)-\omega_{l}|\leq\frac{\Delta\omega}{2}}{\!W_s(a_k,b)a_k^{-3/2}(\Delta a)_k}
\end{equation}
where $\Delta\omega$ presents the width of those frequency bins $[\omega_l-\frac{1}{2}\Delta \omega,\omega_l+\frac{1}{2}\Delta \omega]$, $\Delta\omega = \omega_{l}-\omega_{l-1}$, $(\Delta a)_k=a_{k}-a_{k-1}$ and $T_{s}(\omega_{l},b)$ is the synchrosqueezed transform at the centres $\omega_l$ of the sequential frequency bins. 
Using Equation~(9), at~each fixed time point $b$, the~reassigned frequencies should be estimated for all the scales. 
This can be done by adding up all {contributions incorporating} $W_s(a,b)$ and considering the distance between the candidate IF $\omega_{s}(a,b)$ and $\omega_{l}$ ({see Equation~(9)}). This distance must be within a specified frequency bin width ($\Delta\omega/2$). 

\subsubsection{Inverse Synchrosqueezed Wavelet~Transform}
To obtain the time-frequency spectrum of a desired input signal ($s(t)$) by  SSWT, $T_{s}(\omega_{l},b)$ must be calculated using Equation~(9). 
As an important advantage of SSWT, as shown in~\cite{sst}, the~original signal can be analytically reconstructed succeeding the synchronosqueezing step. 
As expected, $T_{s}(\omega_{l},b)$ is concentrated more sharply around the actual IFs of the original signal (Equation (9)). This is due to the fact that the spectrum derived by the SSWT is more sparse than $W_s(a,b)$ which is obtained by wavelet transform (Equation (5)).

ISSWT is proposed to perform signal reconstruction. Its operation is based on inverting the CWT integrating over the frequencies associated with a desired component having a fully discretized version of Equation~(9) presented as $\tilde{T}_{s}(w_l,t_m)$ and a set of fixed frequency ranges as the input to the ISSWT. These frequency ranges can be specified either by the user or by applying a standard least-squares ridge extraction method~\cite{sst2}. Let us represent these frequencies as $l\in \mathscr{L}(t_m)$, where $m=0,...,n-1$, $t_m=t_0+m\Delta t$, $a_j=2^{j/n_v}\Delta t$, $j=1,...,Ln_v$ and $Ln_v$ is the number of scales. Given the frequencies of a desired component ($k^{th}$), the~corresponding signal can be reconstructed using the following equation:
\begin{equation}
    r_k(t_m)=2\textit{R}^{-1}_{\psi}\Re (\sum_{l \in \mathscr{L}(t_m)}\tilde{T}_{s}(w_l,t_m))
\end{equation}
where $\textit{R}_{\psi}$ as a normalization constant has been defined in~\cite{sst1}. In~this research, we used ISSWT to reconstruct the enhanced SSWT spectrum. To~do this, first the original SSWT was applied to the ECG signal; {then}, the~frequency ridge around HR frequency range was obtained. This ridgewas subjected to ISSWT and then SSWT was applied which is expected less noisy than the original ECG~spectrum. 

\subsubsection{Iterative Pulse/R Peak~Detection}

One key to reliable ECG analysis is to do a robust peak detection. The~proposed method here to extract R peaks of the ECG signals is based on an iterative algorithm. The~iterative algorithm aims to combine the information from R peaks estimated in temporal domain and frequency ridge estimated from the time-frequency spectrum of the ECG. This has been done by comparing the estimated heart rate from the estimated ECG R peaks and frequency ridge with maximum energy around the heart rate frequency. One issue in comparing such HR 
frequencies from these two techniques (time-domain and time-frequency domain) is sampling intervals, and consequently the number of samples, which makes pairwise comparison~difficult. 

To illustrate the difference in sampling intervals, first consider that SSWT has been applied to the ECG signal, and that the frequency ridge (presenting HR frequency) with the highest energy in the range of valid heart rate has been derived. This will produce a new time-series corresponding to instantaneous frequencies. For~this new time-series, the~sampling interval is fixed as 1/fs where fs is the sampling frequency. The~length of this time-series is equal to the number of ECG signal samples.
On the other hand, the~difference in consecutive R peaks is considered to derive a new time-series presenting HR frequency (or heart rate variability). The~sampling interval is simply the difference in the sample numbers between R peaks divided by fs as \hl{diff(R\_peaks\_samples)/fs}.
The length of this time-series is equal to the number of R peaks minus one. 
Suppose there are 30 R peaks detected in a time window of 5000 samples. For~the first time-series presenting HR derived from SSWT, the~length of the time-series is 5000; for~the second one, the~length is 29~samples. 

\textls[-15]{To compare these two time-series, a~uniform sampling is required. For~our experiment, the~maximum frequency} of 2 Hz for heart rate has been considered; therefore, the~resampling of HR frequencies to 4~Hz signal satisfies the Nyquist sampling criterion.
Once both time-series are resampled to a fixed rate, an~absolute error of difference measure can be calculated. Such a measurement of error has been used to indicate the reliability of the R peak detection  algorithm (temporal domain) compared with time-frequency spectral analysis.   
Calculated error might not be zero in the case of having a highly reliable R peak detection method, as~the resampled signals cannot be overlapped thoroughly; however, the~error has been used to tune the peak detection parameter (such as thresholds), and therefore, discard invalid detected R peaks to increase the reliability of and generalize the pulse peak of the detection~algorithm. 

The proposed iterative peak detection algorithm is summarized in Algorithm \ref{Alg1}. The~algorithm includes applying a simple peak detection algorithm with a threshold as input. For~a set of different thresholds (\hl{$\mathbf{t}$)}, the~R peak detection has been applied and two time-series were derived to present HR frequency ($f_1$ from SSWT transform and $f_2$ as from detected R peaks). For~various threshold levels, $f_2$~has been calculated and an absolute difference error of $f_1$ and $f_2$ has been obtained. Finally, a~set of R peaks with minimum error have been selected as the final selected R peaks. If the agreement between two methods resulted in production of a much smaller number of R peaks or a sparse SSWT spectrum, the~ECG segment was marked as noisy with unreliable pulse peaks and not included in the~analysis.


\begin{algorithm}[H]
\caption{Iterative pulse peak detection.}
\label{Alg1}
\begin{itemize}[leftmargin=-5mm,labelsep=0.8mm]
\item[-] \textbf{Input ECG:} $\textbf{s}$ $\leftarrow$ input ECG\\
\item[-] Set sampling frequency $fs$\\
\item[-] Set a vector of thresholds $\mathbf{t}$ for detecting the pulse peaks\\
\item[-] $[W]$ = WSST(\textbf{s},~$fs$) - SSWT transform\\
\item[-] $\mathbf{f}_1$ = TFRIDGE($W$) - frequency ridge in the range of $[0.5-2]$ Hz\\
\item[-] $\hat{\mathbf{f}}_1$= INTERPRR($\mathbf{f}_1$...4...) - Resample frequencies into 4 Hz\\
\item[-] \textbf{for} $thr$ \textbf{in} $\mathbf{t}$\vspace{-8pt}
\begin{itemize}[leftmargin=4.5mm,labelsep=5.8mm]
\item[~] [p] = PEAKDET(\textbf{s},~$thr$)\\
\item[~]$\mathbf{f}_2$ = $fs$/DIFF(p) -  HR frequency\\
\item[~] $\hat{\mathbf{f}}_2$= INTERPRR($\mathbf{f}_2$...4...) - Resample frequencies into 4 Hz\\
\item[~] error($thr$) = ABS($\hat{\mathbf{f}}_2$-$\hat{\mathbf{f}}_1$)\vspace{-4pt}
\end{itemize}
\item[-] Find set of pulse peaks associated with minimum error\vspace{-8pt}
\end{itemize}

\end{algorithm}

\begin{figure}[H]
\begin{center}
\includegraphics[width=140mm,height=110mm]{{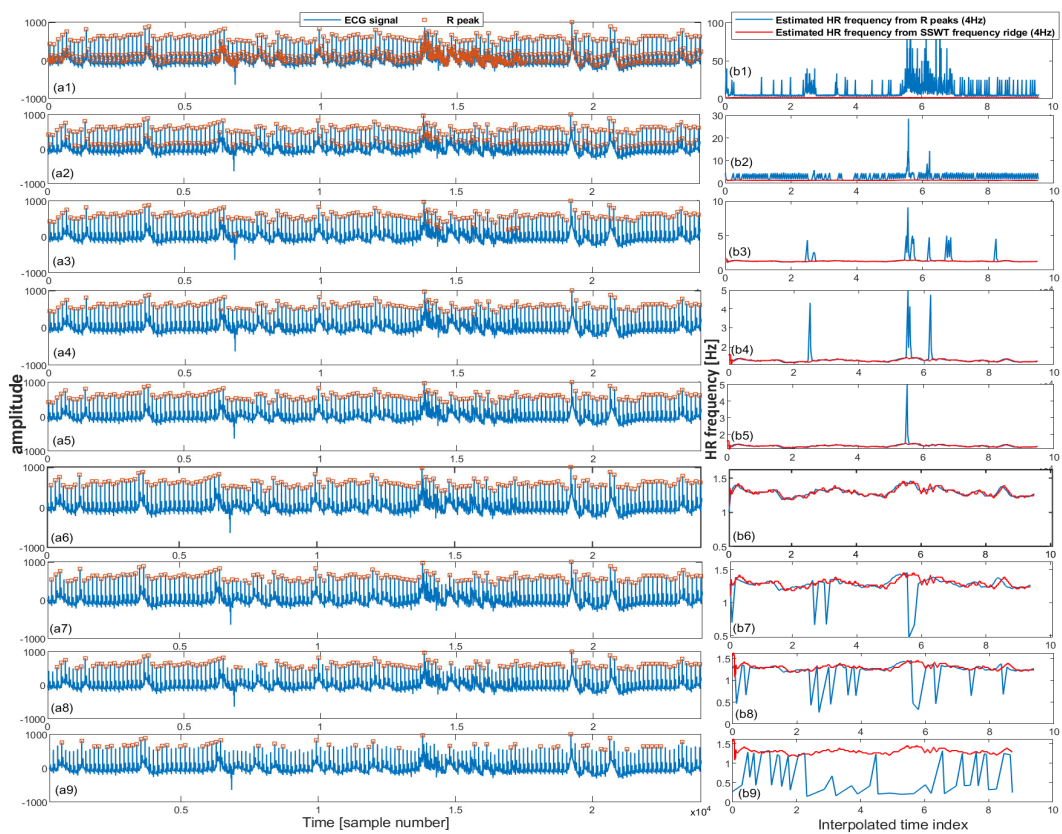}}\\
\caption{\hl{In each row}, selected ECG segments and detected R peaks are shown on the left. In~the right, the~estimated HR frequency from the detected R peaks and \hl{synchrosqueezed wavelet transform (SSWT)} are shown. The threshold \textls[-15]{to detect R peaks increases from top to bottom; (\textbf{b1}--\textbf{b5}) presents overestimated R peaks; (\textbf{b6}) presents the correctly detected R peaks hugely overlapped with the estimated HR 
			frequencies; (\textbf{b7}--\textbf{b9})~presents under estimated R~peaks.}} 
Once R peaks are reliably detected, the~ECG features can be extracted. Here, we have used two types of ECG features. The~first set of features is directly derived from the detected R peaks. The~second set of features is derived from respiratory modulation. The~respiratory modulation is a time-series derived from R peak amplitudes. These two sets are features are explained in the following~subsection. 

\label{fig:sswt}
\end{center}
\end{figure}

\begin{figure}[H]
\begin{center}
\includegraphics[width=125mm,height=75mm]{{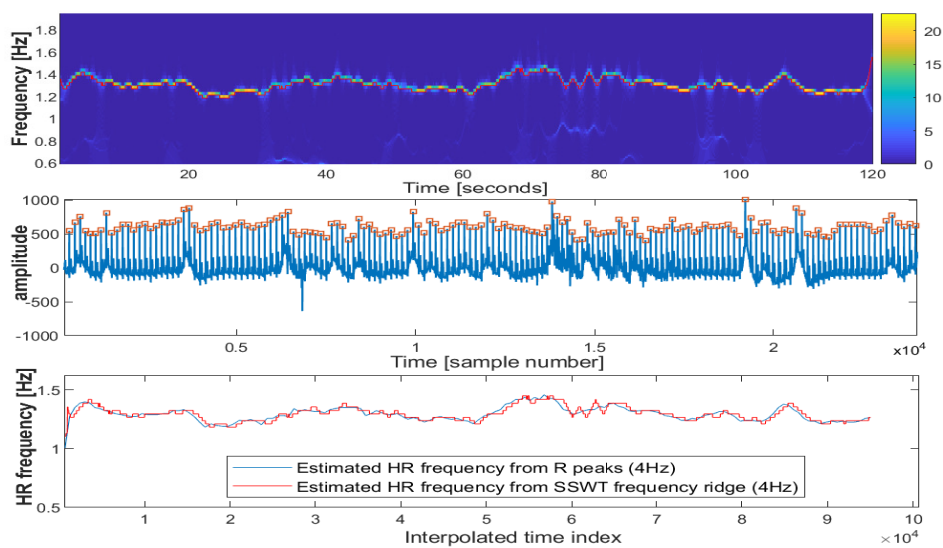}}\\
\caption{{\textbf{Top}}: estimated spectrum using SSWT. \textbf{Middle}: segmented ECG signal. \textbf{Bottom}: estimated HR frequency from R peaks and SSWT (resampled into 4 Hz) overlain.} 
\label{fig:example}
\end{center}
\end{figure}

To further illustrate the algorithm's performance, an~example is shown in Figure~\ref{fig:sswt}. In~this figure, nine different threshold levels have been shown in an increasing order from the top to bottom.
In the right column of Figure~\ref{fig:sswt}, estimated HR frequency from R peaks and SSWT transform are overlaid after resampling into 4 Hz. In Figure~\ref{fig:sswt}b1--b5, overestimated R peaks have created large variations in derived HR frequency, while in Figure~\ref{fig:sswt}b7--b9, there exist underestimated R peaks which have also created large variations. The~plot in Figure~\ref{fig:sswt}b6 corresponds to the minimum error between HR frequency derived from R peaks and SSWT and represents the most reliable detected R peaks on ECG signal. This plot has been shown in Figure~\ref{fig:example} along with its estimated~SSWT.

\vspace{12pt}

\subsubsection{ECG~Features}
ECG features extracted from pulse peaks include maximum, minimum and mean differences of consecutive R peaks. This set of features provides the timing of R peaks and is partly related to heart rate variability and one type of respiratory modulation (frequency modulation) \cite{peter}.
The second set of ECG features is derived from the respiratory modulation. There are three respiratory modulations (frequency, amplitude and intensity) introduced in the literature to be derived from ECG R peaks to present respiratory~\cite{peter,peter1}. Here, we calculated the amplitudes of R peaks to derive respiratory amplitude modulation. Then, maximum, minimum and mean amplitudes of derived time-series were calculated. 
Finally, SSWT was applied to the derived respiratory amplitude and the instantaneous frequencies related to the breathing range [0.05--1] Hz were estimated. Then, maximum, minimum, mean {and standard deviation values of instantaneous} frequencies were estimated. A~summary of ECG features (along with EMG features) is shown in Table~\ref{table:feat}. 


\subsection{A Multimodal Deep Learning Neural Network for Sleep {Disorder} Detection}

\begin{figure}[H]
  \includegraphics[width=\linewidth]{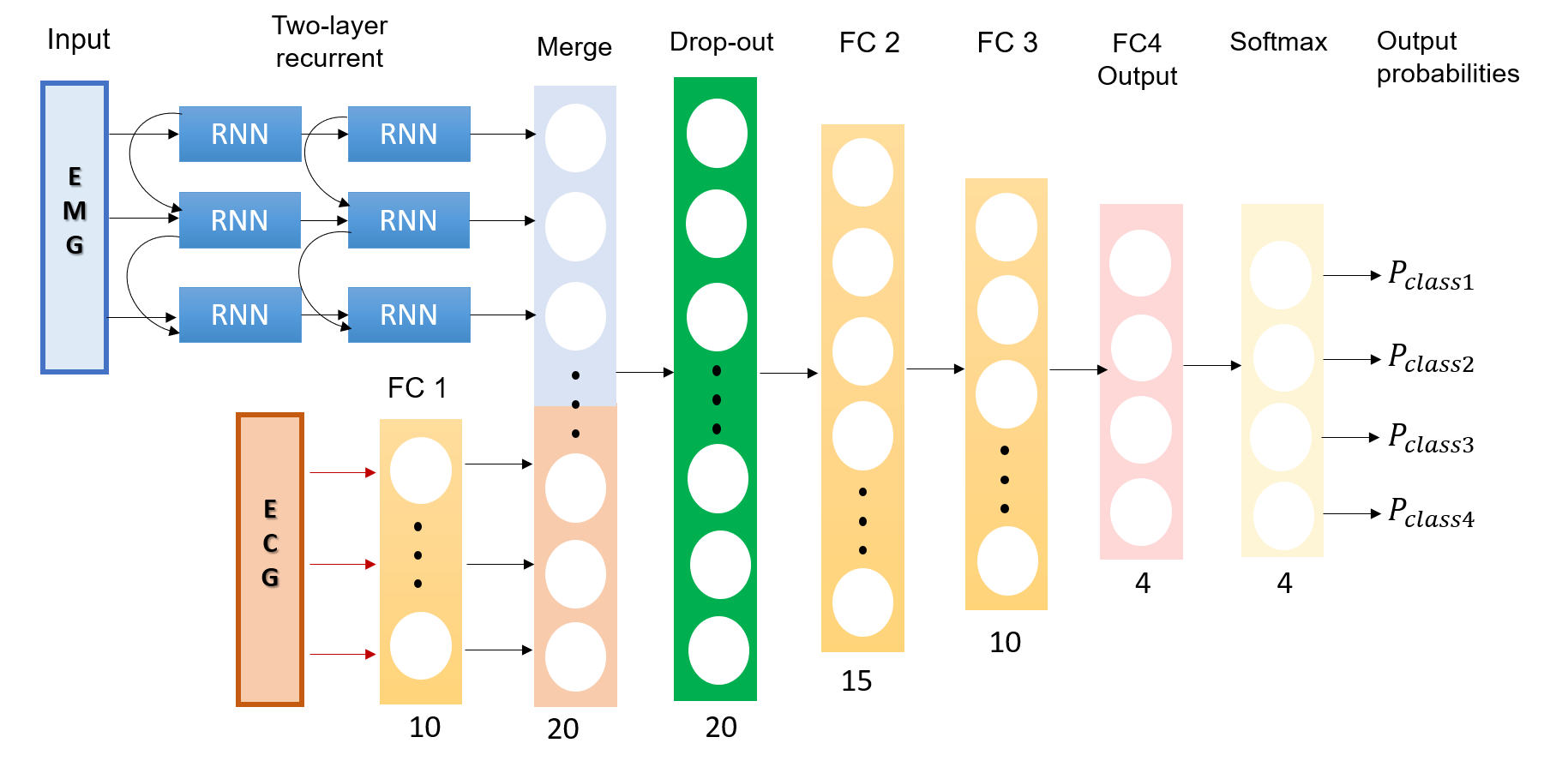}
  \caption{Architecture schema of the proposed multimodal EMG-ECG~DNN.}
  \label{fig:DNN}
\end{figure}

{The EMG can be analyzed by directly extracting the raw data features from signal samples, while ECG should be analysed by applying the proposed techniques in Section~\ref{s2.2} for reliable extraction of pulse peaks followed by derivation of respiratory modulation.   
Then, t}he features extracted from the EMG and ECG (Table \ref{table:feat}) are both used to train a deep learning neural network (DNN).{ Using the proposed DNN, we obtained a performance of 72\% accuracy, 57\% F1 score, 53\% precision and~62\% recall. The~accuracy was significantly better than that of the second best approach test.} A concern when designing the architecture is that the deep neural network should consider that the ECG features summarize a full observational period of a trial (120 s), while for the EMG, each observation represents a sliding window of 2 s each over a trial 24 s. In~terms of deep learning nomenclature, this means that the tensors are of different shapes (dimensions). An~architecture that takes both temporal and structured data tensors and is able to integrate them into the neural network is a desirable option. Our proposed best performing model is a deep learning architecture wherein the temporal variations of the EMG features can be also accommodated. {This is achieved by adding a merging layer} that unites the input layer from the ECG with the output from the recurrent layers that takes the temporal EMG features as inputs, as~shown in Figure~\ref{fig:DNN}. The~layers of deep learning architecture are detailed~as:
\begin{itemize}[leftmargin=*,labelsep=5.8mm]
    \item \textbf{\hl{Input Layer EMG:}}  5\highlight{$\times$}3 
    tensor of the feature layer; the first dimension is the statistical parameters as presented in Table~\ref{table:feat} and the second dimension is temporal window of the trial.
    \item \textbf{\hl{Input Layer ECG:}}  1\highlight{$\times$}10 tensor of statistical parameters of the ECG extracted from Table~\ref{table:feat}.
    \item \textbf{Two hidden RNN {layers} for EMG input:} Two hidden RNN layers with five units each.
    \item \textbf{Fully connected layers for ECG input (FC1):} A fully connected layer \hl{(FC)} that performs a linear transformation with a weight matrix and bias; an ~input length of 10 and a number of hidden units of 20 with a rectifier linear unit (ReLU) activation function.
    \item \textbf{Merging Layer:} A layer that concatenates output tensors from the RNN layers and FC1 as a 1~dimensional tensor 1\highlight{$\times$}25.
    \item \textbf{Drop-out layer:} A drop out layer with attention weights set to 0.4.
    \item \textbf{Fully connected layer (FC1):} A fully connected layer that applies a linear transformation (i.e.,~weight matrix and bias) with 25 inputs and 15 hidden units with ReLU activation.
    \item \textbf{Fully connected layer (FC2):} A fully connected layer that applies a linear transformation with 15~inputs and 15 hidden units with ReLU activation.
    \item \textbf{Fully connected layer (FC3):} A fully connected layer that applies a linear transformation with 15~inputs and 10 hidden units with ReLU activation.
    \item \textbf{Fully connected layer (FC4):} A fully connected layer that applies a linear transformation with 15~inputs and four hidden units.
    \item \textbf{Softmax layer (SF):} This layer re-scales between 0 and 1, with each output tensor element ($x_i$) from the fully connected layer FC4 as $x_{i}=exp(x_{i})\backslash\sum_{j}exp(x_{j})$.
\end{itemize}

The criterion used for this multi-class problem is the negative log likelihood loss. An~adaptive learning rate optimizer (ADAM) is used to train the neural network. {The initial layers of the \hl{DNN} architecture were designed to incorporate the temporal information provided by EMG alongside the ECG information. That is, keeping as much information as possible from the extracted electrophysiology signal markers. Beyond~the merging layer, the~number of \hl{FC} layers has been decided based on a simple hit-or-miss approach. That is, we gradually add layers until no further improvement can be achieved.  
For the selection of the hyper-parameters, including the number of hidden units for each FC layer, the drop out rate and the learning rate for ADAM, we employed a tree-structured Parzen estimator search algorithm (Bergstra, 2011). For~the pruning and early termination of bad trials, we used a custom termination condition that checks that the loss reduction is larger than \hl{10e-3} 
every 100 epochs during the training loop, and otherwise terminates it}.

\section{Results}\label{s3}

{A set of 76 patients including
healthy subjects (33) and patients with
OSA only (25), and~both OSA and RLS (18), was used in~\cite{ales1} for sleep scoring. This dataset was extended to include patients with RLS only. Then, a~subset of this extended dataset was selected using 10 subjects from each group to evaluate our developed framework. Therefore,} in this research, a~dataset of  40 subjects, including a set of healthy subjects (10) and patients with
OSA only (10), RLS only (10) and~both OSA and RLS (10) has been~used.  

One ECG channel and an EMG (attached to leg)
were used to evaluate the proposed framework. The~sampling frequencies of both ECG and EMG signals were fixed at 200 Hz.
{Both ECG and EMG signals are divided into 120 s data segments. Both signals were sliding every 60 s (50\% overlap). The~EMG signal aligned with ECG segment, is further divided into 20 s window size sliding every 2 s (90\% overlap) to extract EMG features. This creates three values for each feature of the EMG signal (five features summarized in Table~\ref{table:feat}) and 15 features for each EMG window. However, the~ECG signal is evaluated across 120 s to extract 10 features, summarized in Table~\ref{table:feat},
in each ECG window.
Due to continuous recording of the signals, those signal segments corrupted with noise and artifacts were discarded. In~total, the~numbers of observations we obtained for the groups were: (4378$\sim$33\%) for healthy subjects, (4590$\sim$34\%) for patients with
OSA only, (2225$\sim$17\%) for RLS only and~(2153$\sim$16\%) for both OSA and RLS.} 

In this section we show the results from the recognition of the four {subject} groups by several state-of-the-art classifiers and the proposed method. The~recognition was performed across subjects (i.e., cross-subject), wherein the training data did not belong to the same subject when the test was performed. This recognition scenario is rather complex because of the subject variability and symptomatic uncertainties; however, it is more desirable in clinical terms for real~applications. 

{\textls[-15]{The methods used for benchmark purposes are a vanilla multi-layer perceptron (MLP) with 15~units} in the hidden layer, in which parameters are trained using back-propagation with the Limited-memory BFGS solver and default settings of scikit-learn~\cite{scikit}; a~support vector machine (SVM) using the libsvm package~\cite{svm} with l2 penalty, hinge loss and multiclass one-vs-rest (OVR); a~linear support vector machine (LSVM) using the liblinear library~\cite{linearsvm} with l2 penalty, hinge loss and OVR; random forest (RF) \cite{randomf} made of 10 estimators and using gini as a split criterion and the  default setup in scikit-learn; a~k-nearest neighbor (KNN) \cite{knn} with number of neighbors set to 5 and default specifications~\cite{scikit}; XGBoost (XGB) using the dmlc XGBoost package~\cite{xgb} and its recommended settings; and~AutoKeras (autok) \cite{autok} an AutoML approach based on neural architecture searching that tunes its architecture and hyper-parameters itself.}

\begin{figure}[H]
\centering
\includegraphics[width=0.96\linewidth]{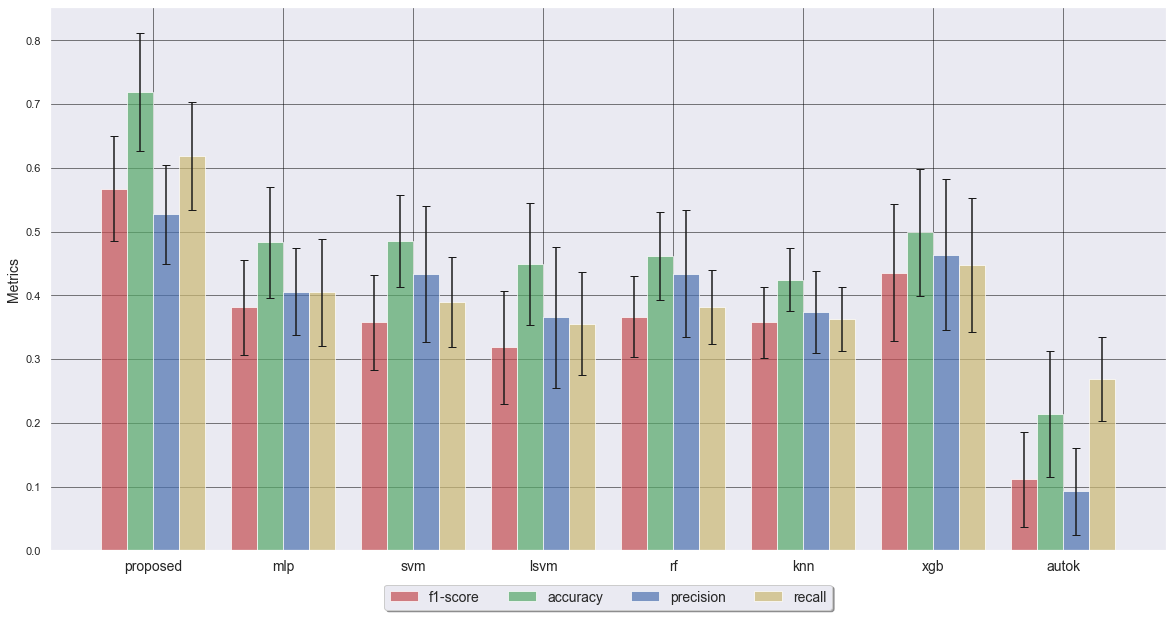}
  \caption{Results comparison of f-scores for the across-subject~model.}
  \label{figcross}
\end{figure}

\textls[-15]{The patient-group recognition was evaluated in a cross-subject recognition scenario for all methods. All the tests were evaluated using 10-fold cross-validation. Examining the results of the cross-subject} recognition scenario (Table~\ref{tab:table2} and Figure~\ref{figcross}) we can see that the performance results are not extremely high but highly promising for such a limited sample size (10 samples for each group only). In~a cross-subject recognition, our proposed multimodal DNN method performance is above most of the other classifiers and it outperforms XGBoost for most subjects (Figure~\ref{figcross}). XGBoost is well known for beating many classifiers in machine learning competitions (such as Kaggle), including deep learning approaches~\cite{niel}. Statistical differences in the performance statistics were found between the proposed DNN method and the rest. It is also worth noting that the AutoML approach tested, AutoKeras~\cite{autok}, that tries to automatically find an optimal DNN architecture, performs very poorly. An~accuracy lower than 80\% in this complex clinical scenario is expected because the symptomatic uncertainty comes into play. Some interesting performance patterns may begin to be appreciated in our study. In~this cross-subject scenario our proposed method significantly outperforms all the others; in~particular, the overall recall of our proposed method differs positively by a wide margin. The~abilities of deep learning for transferring knowledge from one domain to another and integrating variable information are well known. Therefore, our intuition is that if the sample size increases, the~performance of our proposed DNN method will increase with respect to other~methods.

\begin{table}[H]
\centering
\caption{Cross-subject results.} 
\label{tab:table2}
\renewcommand{\arraystretch}{1.3}
\scalebox{0.87}{\begin{tabular}{ l cccccccc}
\toprule
\textbf{Metric} & \textbf{Proposed} & \textbf{MLP} & \textbf{SVM} & \textbf{LSVM} & \textbf{RF} & \textbf{KNN} & \textbf{XGB} & \textbf{AUTOK} \\ \midrule
{F1 Score} & \textbf{\hl{0.57}} $\pm$ 0.08 & 0.38 $\pm$ 0.07 & 0.36 $\pm$ 0.07 & 0.32 $\pm$ 0.09 & 0.37 $\pm$ 0.06 & 0.36 $\pm$ 0.06 & 0.44 $\pm$ 0.11 & 0.11 $\pm$ 0.07 
\\ 
{Accuracy} & \textbf{\hl{0.72}} $\pm$ 0.09 & 0.48 $\pm$ 0.09 & 0.48 $\pm$ 0.07 & 0.45 $\pm$ 0.1 & 0.46 $\pm$ 0.07 & 0.42 $\pm$ 0.05 & 0.5 $\pm$ 0.1 & 0.21 $\pm$ 0.1 \\ 
{Precision} & \textbf{\hl{0.53}} $\pm$ 0.08 & 0.41 $\pm$ 0.07 & 0.43 $\pm$ 0.11 & 0.37 $\pm$ 0.11 & 0.43 $\pm$ 0.1 & 0.37 $\pm$ 0.06 & 0.46 $\pm$ 0.12 & 0.09 $\pm$ 0.07 \\
{Recall} & \textbf{\hl{0.62}} $\pm$ 0.09 & 0.4 $\pm$ 0.08 & 0.39 $\pm$ 0.07 & 0.36 $\pm$ 0.08 & 0.38 $\pm$ 0.06 & 0.36 $\pm$ 0.05 & 0.45 $\pm$ 0.1 & 0.27 $\pm$ 0.07 \\ 

\bottomrule
\end{tabular}}

\end{table}
\unskip
\unskip

\section{Discussion and~Conclusions}\label{s4}

In this research, concurrent analysis of ECG and EMG signals have been used for recognition of breathing and movement-related disorders. 
{Our research} has proposed a set of markers from highly-noisy electrophysiological data that can be useful for the recognition of disease symptoms. Their adequacy was evaluated using a set of different machine learning classifiers. Furthermore, a~DNN architecture that yields competitive and promising results by using and integrating this multimodal data was also presented. The~proposed iterative peak detection algorithm can be applied for processing other signals in various applications, not just for musculoskeletal but for inflammatory disorders too~\cite{ra1}. The~obtained accuracy in this study is very encouraging for continuing the research towards intelligent diagnoses and continuous monitoring of multiple sleep disorders. It should be noted that the accuracy is obtained by considering thousands of observations where some observations do not reflect sleep disorder symptoms, whereas they are included in determination of classification accuracy. In~future work, a~detailed analysis may be performed by accurate identification of sleep disorder symptoms only. However,~this is very difficult, since it involves labeling hours of continuous recording of sensor data. Our method can be used for more detailed data analysis of patients diagnosed with sleep disorder symptoms by further subject-specific analysis of ECG and EMG patterns. 
For future works we aim at expanding the number of bio-signals employed in the recognition. The signal processing techniques developed for ECG analysis can be adapted for the analysis of PPG signals recorded in very unobtrusive~ways. 

\vspace{6pt} 


\authorcontributions{\hl{Conceptualization, D.J., J.A.P., A.P., and S.S.; Methodology, D.J., and J.A.P.; Software, D.J., J.A.P., and M.K.; Formal Analysis, D.J., J.A.P. and M.K.; Data Curation, O.V., and J.K.; Writing – Original Draft Preparation, D.J., J.A.P.; Writing – Review \& Editing, D.J., J.A.P, M.K., O.V., J.K., A.P., and S.S.; Supervision, A.P. and S.S.} All authors have read and agreed to the published version of the manuscript.}
\funding{\hl{There is no funding to be disclosed.}}
\acknowledgments{This research was supported by grant projects of the Ministry of Health of the Czech Republic (FN HK 00179906) and of the Charles University at Prague, Czech Republic (PROGRES Q40), and by the project PERSONMED---Centre for the Development of Personalized Medicine in Age-Related Diseases, regulation number CZ.02.1.01\/0.0\/0.0\/17\_048\/0007441, co-financed by the European Regional Development Fund (ERDF) and the governmental budget of the Czech Republic. No ethical approval was required for this study.}
\conflictsofinterest{The authors declare no conflict of~interest.} 



\appendixtitles{no} 

\reftitle{References}




\end{document}